\documentclass{bmvc2k}

\title{Spatiotemporal Deformable Scene Graphs for Complex Activity Detection}

\addauthor{Salman Khan}{salmankhan@brookes.ac.uk}{1}
\addauthor{Fabio Cuzzolin}{fabio.cuzzolin@brookes.ac.uk}{1}

\addinstitution{
 Visual Artificial Intelligence Laboratory \\
 Oxford Brookes University\\
 Oxford, UK
}

\runninghead{S. khan and F. Cuzzolin}{Complex Activity Detection}

\begin{document}

\maketitle
\vspace{-5mm}
\begin{abstract}
Long-term complex activity recognition and localisation can be crucial for decision making in autonomous systems such as smart cars and surgical robots. 
Here we address the problem 
via a novel deformable, spatiotemporal scene graph approach, consisting of three main building blocks: (i) action tube detection, (ii) the modelling of the deformable geometry of parts, and (iii) a graph convolutional network. Firstly, action tubes are detected in a series of snippets. Next, a new 3D deformable RoI pooling layer is designed for learning the flexible, deformable geometry of the constituent action tubes. Finally, {a scene graph is constructed by considering all parts as nodes and connecting them based on different semantics such as order of appearance, sharing the same action label and feature similarity.} We also contribute fresh temporal complex activity annotation for the recently released ROAD autonomous driving and SARAS-ESAD surgical action datasets 
and show the adaptability of our framework to different domains. 
Our method is shown to significantly outperform graph-based competitors on both augmented datasets.

\end{abstract}

\vspace{-5mm}
\section{Introduction}
\label{sec:intro}

Complex activity recognition is attracting much attention in the computer vision research community due to its significance for a variety of real-world applications, such as autonomous driving \cite{caesar2020nuscenes,camara2020pedestrian}, surveillance \cite{liang2019peeking}, medical robotics \cite{zia2018surgical} or team sports analysis \cite{hu2020progressive}. In autonomous driving, for instance, it is extremely important that the vehicle understands dynamic road scenes, in order, e.g., to accurately predict the intention of pedestrians and forecast their trajectories to inform appropriate decisions. 
In surveillance, group activities rather than actions performed by individuals need to be monitored.
Robotic assistant surgeons need to understand what the main surgeon is doing throughout a complex surgical procedure composed by many short-term actions and events \cite{singh2021saras}, 
in order to suitably assist them. 

\begin{figure*}[h]
    \centering
    \includegraphics[width=0.99\textwidth]{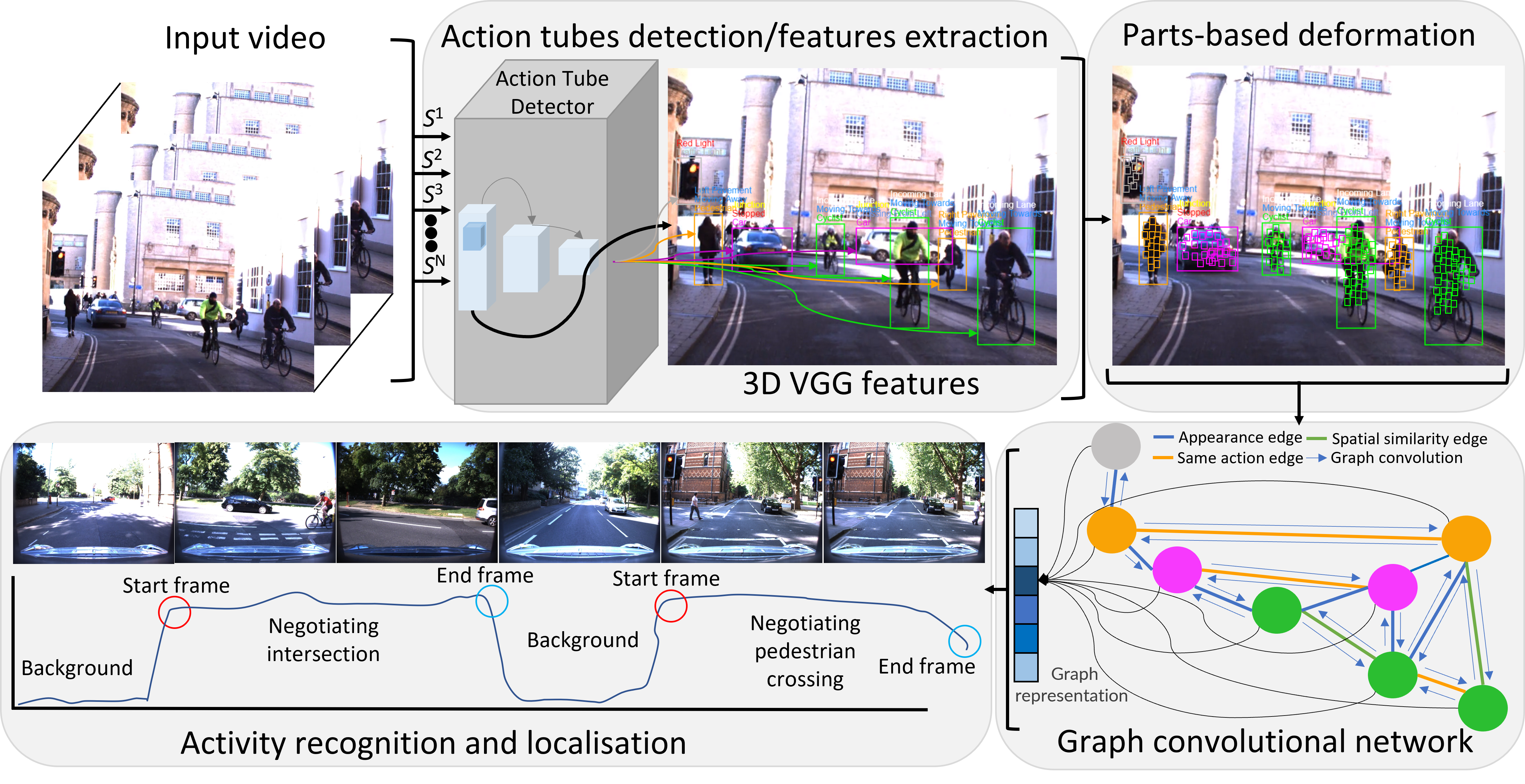}
    \caption{Overall pipeline of our {long-term} complex activity detection framework. (i) The input video is first divided into snippets. (ii) Snippets are passed to an action tube detector module one by one. (iii) A part-based deformation module receives 3D VGG features and action tube locations and returns  features for the salient (non-background) parts of each tube instance. (iv) {A GCN module represents activity parts (action tubes) as nodes with the features generated by ROI pooling and builds edges with different semantics to construct a spatiotemporal graph representation. Finally (v), the graph representation features produced by GCN inference on the consolidated graph are used for temporal activity detection.}} \vspace{-3mm}
    \label{fig:framework}
\end{figure*}

Recent methods for action or activity recognition and localisation 
can be broadly divided into two categories; single atomic action \cite{long2020learning,shi2020weakly,gong2020learning,xu2020g} and multiple atomic action recognition/localisation \cite{huang2020improving,wu2019long,zhao2019hacs,sudhakaran2019lsta,luo2019grouped,ji2020action}. The 
former methods only focus on identifying the start and end of an action performed in a short video portraying a single instance, leveraging datasets such as UCF-101 \cite{soomro2012ucf101} or Charades \cite{sigurdsson2016hollywood}. The latter set of approaches consider videos which contain a number of atomic actions 
or multiple repetitions of the same action. Methods in this category do address complex activity recognition, as their aim is to understand an overall, dynamic scene by detecting and identifying its constituent components. Datasets used for complex activity detection are Epic-kitchens \cite{damen2018scaling}, THUMOS14 \cite{idrees2017thumos} or ActivityNet v1.3 \cite{caba2015activitynet}.
Both classes of methods are geared towards merely recognising and localising short term action or activities that lasts for only a few frames or seconds.

Unlike all existing methods, in this work we present a 
framework capable of recognising \emph{complex, long-term activities}, validated in {the fields of} autonomous driving and surgical robotics 
but of general applicability and extendable to other 
domains.
More precisely, by `complex activity' we mean \emph{an ensemble of `atomic' actions, each performed by an individual agent present in the environment, which extends over a period of time and which collectively has a meaning}.
For instance, 
in the autonomous driving context, one can define the complex activity
\textit{Negotiating intersection} as composed of the following atomic actions: 
an autonomous vehicle (AV) moves along its lane; the traffic light regulating the vehicle's lane turns red, while the light turns green for a traversing road; a number of other vehicles pass through the intersection from both sides; the AV's light turns green again; the AV resumes moving and crosses the intersection.
\\
The proposed pipeline (Fig. \ref{fig:framework}) is divided into three parts: (i) action tube detection, (ii) part-based feature extraction and learning via 3D deformable RoI pooling, and {(iii) a graph generation strategy to process a variable number of parts and their connections, aimed at learning the overall semantics of a dynamic scene representing a complex activity}. Action tube detection is a necessary pre-processing step, aimed at spatially and temporally locating 
the atomic actions present \cite{de2014online,jetley20143d,singh2017online,saha2017amtnet,behl2017incremental,singh2018predicting,saha2016deep}. 
Note that the tube detector needs to ensure a fixed-size representation for each activity part (atomic action). 
Here, in particular, we adopt
AMTNet \cite{saha2020two}, as the latter describes action tubes of any duration using a fixed number of bounding box detections.

Our contribution is twofold. Firstly, our novel 3D deformable RoI pooling layer, inspired by standard deformable and modulated RoI pooling \cite{dai2017deformable,zhu2019deformable}, 
is not only designed to work with 3D data but is also capable of learning feature representations for tubes of variable spatiotemporal shape.
{Secondly, an original Graph Convolutional Network (GCN) module constructs a graph by 
considering individual tubes as nodes and connecting them via edges encoding diverse semantics, namely: appearance order, sharing of the same action label, and spatial feature similarity. The spatiotemporal scene graph so constructed is then processed by a stack of graph convolutional layers resulting in graph representation features, which are used to train a classifier for recognising complex activities, followed by a localisation stage which uses a sliding window approach.}

The framework is evaluated using two {real-world datasets springing from} completely different domains: ROAD \cite{singh2021road} for situation awareness in autonomous driving and SARAS-ESAD \cite{bawa2020esad} for surgical action detection, both providing video-level annotation in the form of (atomic) action tubes. 
In this work we augment these datasets with suitable annotation on the start and end time of each instance of complex activity (road activities in ROAD vs surgical phases in SARAS-ESAD).
The main contributions of this paper are therefore:
\begin{itemize}
\item 
A novel framework for long-term complex activity recognition and localisation. 
\item 
An original deformable 3D RoI pooling approach for flexibly pooling features from the various components of the detected tubes, to create an overall representation for activity parts.
\vspace{-1mm}
\item 
{A spatiotemporal scene graph generation and processing mechanism able to cope with a variable number of parts while learning the overall semantics of an activity class.}
\vspace{-1mm}
\item 
\vspace{-1mm}
Augmented annotation for two newly-released datasets 
aimed at making them suitable benchmarks for future work on complex activity detection.
\end{itemize}
{Our results clearly indicate that the detection task (both at atomic and complex activity level) is extremely challenging on the real-world data which forms these newly annotated benchmarks, when compared to existing academic datasets. We hope this will stimulate further original thinking to address these challenges}.
Our method is shown to clearly outperform two recent state of the art graph-based competitors \cite{zeng2019graph,xu2020g} on both augmented datasets.

\vspace{-4mm}

\section{Related Work}
\vspace{-2mm}
\textbf{Complex activity recognition}. Most recent work on complex activity recognition concerns scalar sensors \cite{bharti2018human,thakur2019improved,zhou2020deep} or combination of both scalar and vision sensors \cite{arzani2020switching,kwon2020imutube}. Recently, though, several vision-based complex activity recognition methods have been proposed \cite{huang2020improving,wu2019long,zhao2019hacs,sudhakaran2019lsta,luo2019grouped,ji2020action} with the goal of understanding an overall scene by recognising and segmenting atomic actions. 
These methods can be further divided into (i) sliding windows approaches \cite{wang2014action,shou2016temporal}, in which an activity classifier is applied to each snippet, and (ii) boundaries analyses \cite{xu2017r,gao2017turn}, in which a model is trained to identify the start and end time of each action.  
Overall, current activity recognition methods are geared to recognise short-term activities via a combination of small atomic actions.

Unlike existing approaches, our objective is to understand \emph{long-term} activities in dynamic scenes, such as 
the \emph{phases} a surgical procedure is broken into, 
whose detection is crucial to inform the decision making of automated robotic assistants.



\textbf{Deformable parts-based models}. Deformable part-based models have been used by the research community for more than a decade \cite{4587597,felzenszwalb2008discriminatively,felzenszwalb2010cascade,hsieh2010segmentation} for object detection and segmentation. 
Following the rapid development of Convolution Neural Networks (CNNs), Girshick et al. \cite{girshick2015deformable} first recognised that deformable part-based models can be implemented for object detection in a CNN formulation, in which each convolution pyramid is fed to a distance transform pooling and a geometric filter layer. The main limitation of this method is that it is not end-to-end trainable and requires a heuristic selection of part sizes and components. A subsequent end-to-end deformable CNN formulation was proposed in \cite{dai2017deformable}, which uses two new CNN layers (deformable convolution and deformable RoI pooling) that reproduce the functionalities of traditional part deformation. The latest version of deformable CNN is Deformable ConvNets v2 \cite{zhu2019deformable}, which introduces a modulation mechanism in both deformable convolution and RoI pooling.

To the best of our knowledge, all deformable models proposed to date focus on either object or short-term action detection, whereas here, {for the first time, we design a novel 3D deformable RoI pooling layer for learning \emph{long-term} complex activities.} 

\textbf{Graph convolutional network}. { Recently, GCNs have been widely used for action and activity detection and recognition, building on their success in different areas of computer vision such as point cloud segmentation \cite{wang2019dynamic,xie2020point} and 3D object detection \cite{gkioxari2019mesh}. Relevant GCN approaches have been broadly focussing on either action recognition \cite{wang2018videos,liu2019learning,chen2019graph} or temporal action localisation \cite{zeng2019graph,xu2020g}. 
In the former, videos are represented in different spatiotemporal formats such as 3D point clouds and time-space region graphs, {and methods focus on recognising atomic actions only.}
In contrast, Zeng et al. \cite{zeng2019graph} use GCN for temporal activity localisation by considering action proposals as nodes and a relation between two proposals as an edge. {In opposition, in our model nodes are action tubes and their connections are based on an array of semantics.}
In another recent study, Xu et al. \cite{xu2020g} generate graphs by considering temporal snippets as nodes and drawing connections between them based on temporal appearance and semantic similarity.} 

{Most graph-based activity detection methods \cite{xu2020g,li2020graph,zeng2019graph} construct a graph for a whole video by taking snippets as nodes and their temporal linkage as edges, not paying much attention to the constituent atomic actions within each snippet, and are typically limited to shorter videos and memory dependent. In contrast, our proposed framework is designed to construct a graph for each snippet which reflects the structure of a dynamic scene in terms of atomic action tubes (nodes) and the different types of relationships between them.}

\vspace{-4mm}

\section{Proposed Method}

Crucial to the identification of complex video activities is 
the modelling of the relations among the constituent actions. 
In this paper we propose to achieve this via a combination of the deformable pooling of features and a spatiotemporal graph representation employing multiple semantics.

\vspace{-4mm}
\subsection{Action Tube Detection}

To provide a fixed-size representation for the instances of atomic actions composing a complex activity,
here we adopt AMTnet \cite{saha2020two}. AMTnet is a two-stream online action tube detector that uses both RGB and optical flow information (although here we only use the RGB stream). 
The main rationale for using AMTnet is that it generates tubes in an incremental manner while preserving a fixed-size representation. 

\textbf{Architectural Details}. AMTNet uses VGG-16 \cite{simonyan2014very} as baseline CNN feature extractor. The last two fully-connected layers of VGG-16 are replaced by two convolutional layers. Four extra convolutional layers are added at the end. 
AMTNet takes as input a sequence of RGB frames  with a fixed temporal interval $\triangle$ between consecutive frames, i.e., $\{ {f}_t, {f}_{t + \triangle} \}$. The input to AMTNet is in the format [$BS \times {Sq} \times {D} \times {H} \times {W}$], where ${BS}$ is the training batch size, ${Sq}$ is the sequence length (in this case a pair), $D$ is the dimensionality (equal to 3 as we are dealing with RGB frames), while ${H}$ and ${W}$ are the height and width of each frame ($300 \times 300$ in our case). As typical in action detection, AMTNet uses both a classification and a regression layer for recognition and detection, respectively, with the goal of predicting action `micro-tubes' defined by pairs of consecutive detections. 
The method predicts bounding boxes for a pair of frames separated by fixed gap $\triangle$, while the bounding boxes for intermediate frames are generated by interpolation. In this work, atomic action instances are represented as 3 micro-tubes with $\triangle$ = 3 for an overall tube length of $L = 12$ frames, aligned with our snippet length.
Complete action tubes are incrementally generated by AMTNet by temporally linking the micro-tubes predicted by the network \cite{saha2020two}. 



\subsection{3D Part-Based Deformable Layer}

The feature extractor in our framework is a novel \emph{3D deformable RoI pooling layer} encoding the spatiotemporal geometry of the action tubes which correspond to the activity parts. This is an extension of the existing standard deformable RoI pooling layer \cite{dai2017deformable}, and has the ability to extract and learn features from an action tube rather than a 2D bounding box. 
{The rationale behind using the 3D deformable RoI pooling layer is that it allow us to learn at training time how the 
geometric shape of atomic action tubes (as regions of the video considered as a spatiotemporal volume) varies across instances of the same class.
Intuitively, the shape of the bounding box detections forming a tube (and therefore the shape of the tube itself) will vary with, e.g., the viewpoint, as well as the particular style with which the action is performed by a certain agent (for instance, a cyclist can turn right making a narrower as opposed to a wider turn).} 

\begin{figure}
\centering
\begin{minipage}{.52\textwidth}
    \centering
    \includegraphics[width=0.97\textwidth]{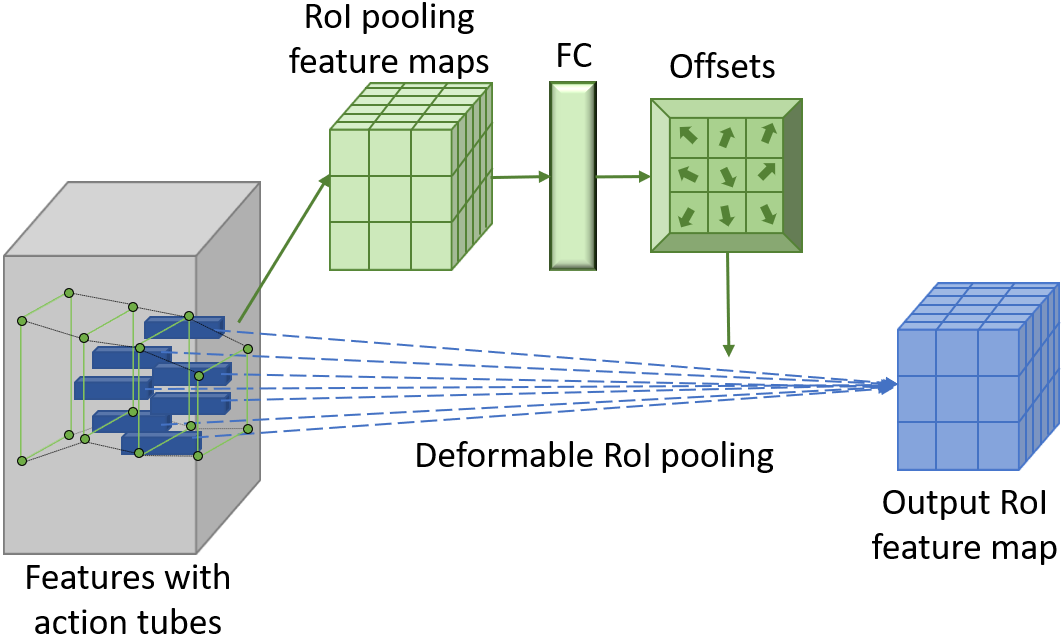}
    \caption{Our 3D deformable RoI pooling layer takes feature maps and action tube locations as input and arranges them into a fixed-size grid of components (here illustrated for size $3\times 3$). For each grid component an offset is generated and multiplied by the original tube feature to produce the final component features.}
    \label{fig:droi}
\end{minipage}%
\hspace{0.1cm}
\begin{minipage}{.46\textwidth}
    \centering
    \includegraphics[width=0.98\textwidth]{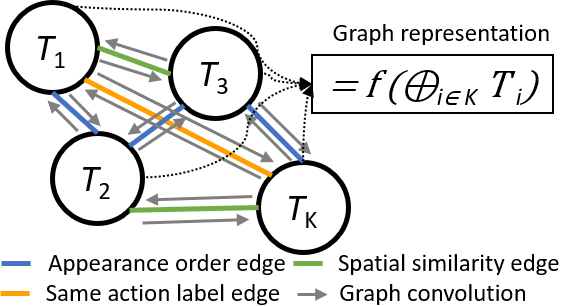}
    \caption{Our graph module takes as input the features generated for each tube by the ROI pooling layer $T_{1,2,3...K}$ and builds edges between them according to different semantics ({order of appearance, spatial similarity, label}). 
    The overall graph is processed by a GCN to deliver a fixed-size graph representation.}
    \label{fig:gcn}
\end{minipage} \vspace{-6mm}
\end{figure}

The principle of our 3D deformable RoI pooling operation is shown in Figure ~\ref{fig:droi}. Like the classical deformable RoI pooling layer, our module also includes standard RoI pooling (used in all region proposal-based object detection methods), a fully connected layer, and {\emph{offsets} which encode the amount of geometric deformation. 
Firstly, standard RoI pooling is applied to the provided feature map $X$ and bounding box locations forming an action tube ($L$ $\times$ [$x$,$y$,$w$,$h$]), by subdividing the tube into a pooled feature map grid of fixed-size in both the spatial and the temporal dimensions: $L$ $\times$ $k$ $\times$ $k$. Here $L = 12$ is the fixed action tube length,
while $k$ is a free parameter which determines the `bin size', i.e., the number of grid locations detections are divided into in each spatial dimension (see Figure \ref{fig:droi} again). 
Next, for each bin in the grid, normalised offsets (representing the degree of deformation of the grid components of each action tube) are generated for these feature maps using a fully-connected layer, which are then transformed using an element-wise product with the original RoI's width and height. Offsets are also multiplied by a scalar value to modulate their magnitude (empirically set to 0.1), making them invariant to the different possible sizes of the RoI. In our framework, this layer takes the VGG features extracted by AMTNet and each detected action tube separately as an input, and returns an overall feature map which encodes both the appearance and the shape (through the above offsets) of each atomic action.}

\vspace{-4mm}
\subsection{Graph Convolutional Network}

As our purpose is to achieve a comprehensive understanding of the dynamic scene which comprises a complex activity,
we propose to use a graph convolutional network to model and exploit the relations between the constituent action tubes. 
Unlike the tree structure of classical part-based models (which requires to fix the number of parts \cite{chen2018part}), 
(spatiotemporal) graphs allow us to flexibly describe a complex activity composed by a variable number of actions (nodes) of different type, and to encode the different semantic relationships between them.
The functioning of our GCN module is illustrated in Figure ~\ref{fig:gcn}.

\textbf{Graph Construction}. 
{When constructing the activity graphs, the input video is subdivided into consecutive, non-overlapping, fixed-length snippets.}
For each snippet, a separate graph is built with a variable number of nodes corresponding to the number of detected activity parts (action tubes) { within the snippet}. The initial representation of the nodes is provided by their RoI features. 
We consider three different types of connections: (i) the \emph{order of appearance} (from left to right) of each action tube (bearing in mind that in autonomous driving, for instance, road activities tend to follow a specific order, e.g., pedestrian crossing the road followed by vehicles engaging an intersection); (ii) the \emph{spatial similarity} of node features, measured using the distance proposed in \cite{wang2019dynamic}; (iii) \emph{node type}, meant as the 
sharing of the same action label, as this provides relevant information for the determination of the activity class. 
{As a result, 
three spatiotemporal scene graphs are constructed, having the same nodes but with different edges. While the second and third graph are undirected, the appearance order graph is a directed one. However, when merging the three graphs an undirected version of the order graph is used. These graphs are then combined by taking a union of all edges to create a single homogeneous graph representing the overall scene. Namely, two nodes are connected in the merged graph iff they are connected in at least one of the three graphs.} 

\textbf{Graph Convolution and Representation}. 
Given the final graph, global graph embedding is applied to extract the context of each snippet portraying a complex activity. In our GCN approach we apply a stack of three graph convolutional layers followed by a graph readout layer. 
The latter encapsulates the final graph representation by taking the mean of the hidden convolutional representations, resulting in fixed-sized feature vectors which are invariant to the number of nodes and edges. 

\vspace{-4mm}

\subsection{Complete Framework}

The complete framework is the concatenation of the aforementioned three modules. Firstly, we divide the video ${V}$ into ${N}$ snippets ${S}_{1,2,3,...,N}$, with each snippet ${S}_{i}$ consisting of a fixed (${M}$) number of frames: $S_i = {F}^{1,2,3,...,M}$. Each snippet is passed to the action tube detection module ${AT}$ which returns ${K}$ action tubes each composed by $L = 12$ bounding boxes with labels ${B}_{A}$ and intermediate VGG features ${X}_{A}$, represented as ${B_A}^{M\times K\times5}$, ${X_A}^{M\times 64\times 300\times300 }$ $\in$ ${AT}$. 
Action tube locations 
and features are then passed to our 3D deformable RoI pooling layer ${DRoI}$ which returns a fixed-sized (i.e., $7 \times 7$) grid of components 
whose dimensionality is equal to the number (64) of convolutional layers: ${{X}_{DRoI}}^{K\times 64\times M\times 7\times7 }$. 
{ These features are then fed to the GCN module ${G}$, where a graph with $K$ nodes and $E$ edges is processed to yield a fixed-sized feature representation ${X_{G}}^{2048}$ $\in$ ${G}$. 
Finally, the latter features are fed to a softmax classifier to classify the snippet into their respective activity category.}

For localisation, we use a sliding window approach {in which snippets are sequentially classified, using a dual verification mechanism. Namely, as our purpose is to recognise and detect \emph{long-term} activities, if there appears to be a random false positive or false negative between two snippets belonging to the same class we simply ignore it and consider it as having the same activity label. The detection algorithm is described in detail in the \textbf{Supplementary material}, Algorithm 1.} 

\textbf{Implementation}. Before training our overall architecture, we separately train AMTNet for action tube detection over both datasets. 
Note that we had to design from scratch suitable data loaders for the two datasets, as 
the format of the annotation there is completely different from that of the original datasets AMTNet was validated upon.
As mentioned, our 3D RoI pooling layer includes a temporal dimension to learn 
the deformation of the components of a tube, rather than of a 2D object.
In our experiments we also adapt a more recent version of deformable RoI pooling, termed \emph{modulated} deformable RoI pooling, to the 3D case. 
{In the GCN module, we construct a graph for each snippet in an online fashion at training time using a PyTorch data loader \cite{NEURIPS2019_9015}. For the design of the GCN architecture we used the Deep Graph Library (DGL) \cite{wang2019dgl} with a PyTorch back-end, which supports the processing of graphs of various length in a single mini-batch.} Overall, our architecture is implemented using the PyTorch deep learning library \cite{NEURIPS2019_9015} with OpenCV and Scikit-learn. For training we used a machine equipped with 4 Nvidia GTX 1080 GPUs with 12GB VRAM each.

\vspace{-5mm}
\section{Experimental Results}
\vspace{-3mm}
\subsection{Datasets and Evaluation Metrics}

In this paper we used two datasets for evaluating our approach, 
both already annotated at video level for action tubes detection. 

\textbf{ROAD} \cite{singh2021road}: ROAD (the ROad event Awareness Dataset for autonomous driving) is annotated for road action and event detection. Each event is described in terms of three different labels: (road) agent (e.g., cyclist, bus), action performed by the agent (e.g., turning left, right), and event location (w.r.t. the autonomous vehicle). The ROAD dataset consists of total 22 videos carefully selected from the Oxford RobotCar Dataset because of their diverse weather and lighting conditions. ROAD comprises 560K bounding boxes in 122K annotated frames with 
560K agent labels, 640K action labels and 
499K location labels. 

For this work we augmented the annotation of the ROAD dataset for complex road activity detection. We used a total of 19 videos with an average duration of 8 minutes each, 12 of which were selected for training and the remaining 7 for testing. We temporally annotated the ROAD videos by specifying the start and end frame for six different classes of complex road activities we inferred from video inspection. For example, a `Negotiating intersection' activity class can be defined which is made up of the following `atomic’ events: Autonomous Vehicle (AV)-move + Vehicle traffic light / Green + AV-stop + Vehicle(s) / Stopped / At junction+ AV-move. {Activity class statistics are listed and described in more detail in the \textbf{Supplementary material}}.  

\textbf{SARAS-ESAD} \cite{bawa2020esad}: ESAD (the Endoscopic Surgeon Action Detection Dataset) is a benchmark devised for surgeon action detection in real-world endoscopic surgery videos. In ESAD, surgeon actions are classed into 21 different categories and annotated with the help of professional surgeons. 
Here we took a step forward and annotated ESAD 
in terms of complex activities corresponding to the different \emph{phases} of the surgical procedure portrayed by the videos (namely, radical prostatectomy). For example, Phase \# 3 corresponds to `Bladder neck transection', in which a scissor cuts the neck of the bladder until it is transected. {Phases and their statistics are again reported in the \textbf{Supplementary material}}. 
{The complex activities (surgical phases) in SARAS were defined by professional surgeons experts in radical prostatectomy. As standard in the surgical context, such phases are consecutive without the need for any background activity class.}
For more details please see \cite{bawa2020esad}.

\textbf{Evaluation Metrics}: For the evaluation of action tube detection performance we used the standard frame/video mean Average Precision (\textit{mAP}) at different IoU thresholds $\delta$ (namely, 0.2, 0.3, 0.5, 0.75) on both datasets. Complex activity recognition was evaluated using classification accuracy, precision, recall and F-score. For complex activity localisation we used the standard protocol \textit{mAP} over the temporal dimension used by all relevant methods.

\vspace{-4mm}

\subsection{Action Tube Detection}

A detailed comparative analysis of AMTNet over different action detection datasets can be found in the original paper \cite{saha2020two}. Here 
we briefly report the performance of AMTNet on our two datasets of choice, as 
AMTNet was never tested there. 
Table \ref{tab:tab1} reports both frame-\textit{mAP} and video-\textit{mAP} results, and compares AMTNet with the proposed baselines 
for the two datasets: the ROAD baseline (termed \emph{3D RetinaNet} \cite{singh2021road}), and 
the ESAD baseline \cite{bawa2020esad}, a vanilla implementation of RetinaNet (only providing frame-level results). 
AMTNet performed better than \cite{bawa2020esad} on SARAS-ESAD, while being inferior to \cite{singh2021road} on ROAD. Remember that the main rationale for using AMTNet is that it can provide a fixed-size representation for the tubes (as required by our framework), motivating us to compromise on accuracy.

\vspace{-4mm}

\subsection{Complex Activity Recognition}

\begin{figure*}[h]
    \centering
    \includegraphics[width=0.98\textwidth]{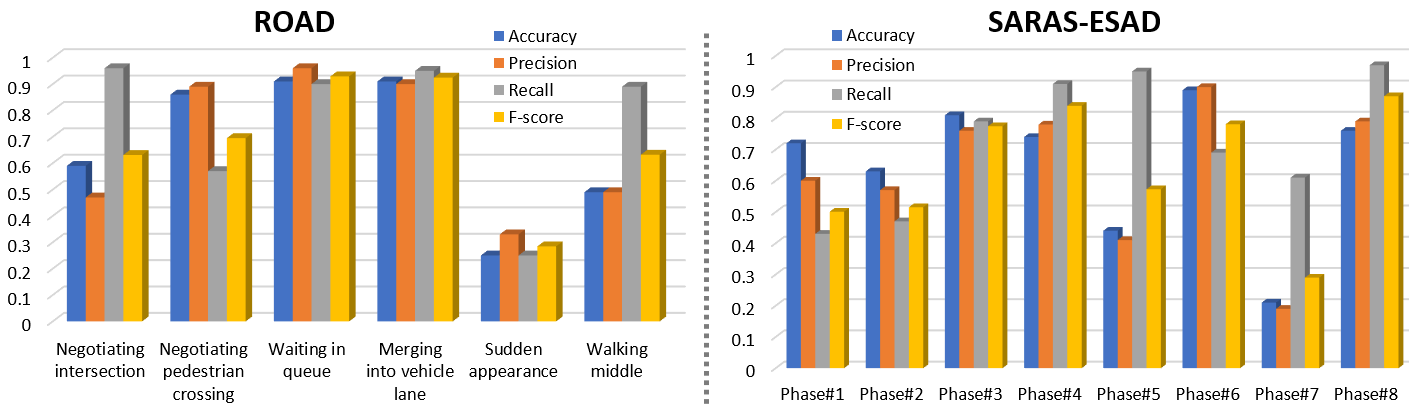}
    \caption{Complex activity classification performance on both ROAD and SARAS-ESAD.\vspace{-2mm}
    }
    \label{fig:reg_res}
\end{figure*}

Next, we provide a detailed analysis of complex activity \emph{classification} using our approach on both the ROAD and SARAS-ESAD datasets. The performance for each class in both datasets is illustrated in Fig. ~\ref{fig:reg_res} using all metrics. 
It is apparent that the ROAD dataset is characterised by significant fluctuations in class-wise performance, with higher recognition accuracy for activities that appear more often, e.g. `waiting in a queue', as opposed to infrequent ones (e.g. `sudden appearance'). 
In SARAS-ESAD each activity class does contain enough samples for good training, while the diversity in phases 
still poses a challenge. 
{ In fact, performance on this dataset is a function of the complexity of the phases and the visual similarity of the constituent atomic actions, not just the amount training data. For example, Phase 5 and 6 both include ‘fat removal’ actions, complicating their differentiation.}

\begin{table*}
\begin{center}
\footnotesize \scalebox{1} {
\begin{tabular}{*{5}{p{0.050\textwidth}}|*{4}{p{0.050\textwidth}}|*{4}{p{0.050\textwidth}}}
\hline
 \multicolumn{5}{c}{}& \multicolumn{4}{c|}{ROAD}  & \multicolumn{4}{c}{SARAS-ESAD}\\
\hline
\multicolumn{5}{c|}{Methods / IoU threshold $\delta$}  & 0.2 & 0.3 & 0.5 & 0.75  &  0.2 &0.3 &0.5 & 0.75\\
\hline
\multicolumn{5}{c|}{Singh et al. \cite{singh2021road} (frame-\textit{mAP}) }  & - & - & 25.9 & -&  -&- &-&-   \\
\multicolumn{5}{c|}{Singh et al. \cite{singh2021road} (video-\textit{mAP}) }  & 17.5 & - & 4.6 & - &  -&- &-&-   \\
\multicolumn{5}{c|}{Bawa et al. \cite{bawa2020esad} (frame-\textit{mAP}) }  & -& - & - & - &  - &24.3 & 12.2 & -  \\
\multicolumn{5}{c|}{AMTNet (frame-\textit{mAP})}  & 22.3 & 18.1 & 15.4 & 11.0 &   30.4 &24.6 &18.7&7.9   \\
\multicolumn{5}{c|}{AMTNet (video-\textit{mAP})}  & 11.6 & 7.9 & 3.8 & - & 13.7 & 10.1 &8.8 &5.4  \\

\hline
\end{tabular}
}
\end{center}
\caption{Action tube detection performance on both the ROAD and SARAS-ESAD datasets. Both Frame-\textit{mAP} and Video-\textit{mAP} at different IoU thresholds are reported for evaluation.\vspace{-2mm}}
\label{tab:tab1}
\end{table*}

\vspace{-4mm}
\subsection{Temporal Activity Detection - Comparison with State-of-the-Art}

To evaluate the performance of our complex activity detection approach we reimplemented two state-of-the-art activity localisation methods -- P-GCN \cite{zeng2019graph} and G-TAD \cite{xu2020g}, as 
neither ROAD nor ESAD were ever used for
complex activity detection. The major changes we made during re-implementation are: (i) data loading (as both P-GCN and G-TAD were designed to be trained and tested on pre-extracted features), and (ii) replacing the regression part with a sliding window approach for the SARAS-ESAD dataset, as the latter lacks a background class.

{\textbf{ROAD}. For activity detection in ROAD we used an additional `background' class, which indicates either no action or presence of action(s) without any solid indication. 
Whenever no action tube was detected we would use the entire frame 
as RoI for our parts deformation module to understand the overall scene. 
Temporal activity detection performance on ROAD, measured via \textit{mAP} at five different IoU thresholds, is reported in Table ~\ref{tab:tab2} for both our approach and the two competitors. Class-wise results for each complex activity 
at a standard IoU threshold of 0.5 are reported in Table ~\ref{tab:tab3}.   }

{\textbf{SARAS-ESAD}. Temporal activity detection on this dataset much relates to 
activity recognition, as surgical phases are contiguous. 
Both the average \textit{mAP} of the methods at five IoU thresholds and the class-wise performance for each activity (phase) at a standard IoU threshold of 0.5 are reported in Table~\ref{tab:tab2} and \ref{tab:tab4}, respectively. From the results it is clear how our method outperforms the chosen state-of-the-art methods by a reasonable margin.}

\begin{table*}
\begin{center}
\footnotesize \scalebox{1} {
\begin{tabular}{*{5}{p{0.038\textwidth}}|*{5}{p{0.038\textwidth}}|*{5}{p{0.038\textwidth}}}
\hline
 \multicolumn{5}{c}{}& \multicolumn{5}{c|}{ROAD}  & \multicolumn{5}{c}{SARAS-ESAD}\\
\hline
\multicolumn{5}{c|}{Methods / IoU threshold $\delta$}  & 0.1 & 0.2 & 0.3 & 0.4& 0.5  &  0.1 &0.2 &0.3 & 0.4 & 0.5\\
\hline
\multicolumn{5}{c|}{P-GCN \cite{zeng2019graph} }  &60.0 & 56.7 & 53.9 & 50.5 & 46.4& 57.9 & 55.6 & 53.4 & 49.0 & 45.1   \\
\multicolumn{5}{c|}{G-TAD \cite{xu2020g} }  &62.1 & 59.8 & 55.6 & 52.2 & 48.7& 59.1 & 56.7 & 54.5 & 49.8 & 46.9 \\
\multicolumn{5}{c|}{\textbf{Ours} }  &\textbf{77.3} & \textbf{74.6} & \textbf{71.2} & \textbf{66.7} & \textbf{59.4}& \textbf{62.9} & \textbf{59.6} & \textbf{58.2} & \textbf{55.3} & \textbf{51.5}  \\

\hline
\end{tabular}
}
\end{center}
\caption{{Comparative analysis of temporal activity localisation performance on ROAD and SARAS-ESAD, 
reporting \textit{mAP} (\%) at five different IoU thresholds.}}
\label{tab:tab2}
\end{table*}

\begin{table*}
\begin{center}
\footnotesize \scalebox{1} {
\begin{tabular}{*{1}{p{0.11\textwidth}}|*{6}{p{0.11\textwidth}}}
\hline
Method / Activities &  Negotiating intersection &Negotiating pedestrian crossing& Waiting in queue &Merging into vehicle lane& Sudden appearance & Walking middle of road\\
\hline
P-GCN \cite{zeng2019graph}  & 44.3 &	53.8	&74.4&	50.1&	21.7&	34.1\\
G-TAD \cite{xu2020g}  & 47.8&	57.3&	70.6&	55.2&	\textbf{24.3}&	37.1
  \\
\textbf{Ours} & \textbf{51.2}&	\textbf{72.3}&	\textbf{89.8}&	\textbf{84.1}&	17.8&	\textbf{41.3}  \\
\hline
\end{tabular}
}
\end{center}
\caption{{ROAD activity localisation performance (\textit{mAP}, \%) for each complex road activity, at a standard IoU threshold of 0.5.} \vspace{-2mm}}
\label{tab:tab3}
\end{table*}

\begin{table*}
\begin{center}
\footnotesize \scalebox{1} {
\begin{tabular}{*{10}{p{0.065\textwidth}}}
\hline
\multicolumn{2}{c|}{Method / Activities} & Phase\#1 & Phase\#2& Phase\#3& Phase\#4& Phase\#5 & Phase\#6 & Phase\#7 & Phase\#8\\
\hline
\multicolumn{2}{c|}{P-GCN \cite{zeng2019graph}} & 56.7&	43.2&	52.3&	59.1&	\textbf{33.8}&	59.4&	14.8&	41.2 \\
\multicolumn{2}{c|}{G-TAD \cite{xu2020g}} & 51.1&	46.6&	57.2	&63.8	&29.4	&62.2&	\textbf{19.3}&	45.7\\
\multicolumn{2}{c|}{\textbf{Ours}}& \textbf{57.5}&	\textbf{54.1}&	\textbf{69.3}&	\textbf{60.2}&	31.1&	\textbf{71.3}&	16.5&	\textbf{52.4} \\

\hline
\end{tabular}
}
\end{center}
\caption{{SARAS-ESAD activity localisation performance (\textit{mAP}, \%) for each activity at a fixed IoU threshold of 0.5.}}
\label{tab:tab4}
\end{table*}
   
\vspace{-4mm}

\subsection{Limitations and Future Work}

The main limitation of this work is that it relies on action tube detection. From our results, the existing tube detectors are not reliable enough to perform well over challenging real-world datasets such as those we adopted here. 
Clearly, if the tube detector misses an important 
atomic action this will affect the overall activity detection performance. 
Nevertheless our results show that, even when using a suboptimal detector, our approach is capable of significantly outperforming state of the art methods on our new benchmarks.

{
Detection is challenging on SARAS and ROAD because of their real-world nature: surgical images are indistinct, road scenes come with incredible variations. These benchmarks show how even the best detectors suffer a huge drop in performance when moving from ‘academic’ benchmarks to real ones. We hope the realism of these two extremely challenging datasets will stimulate real progress and new original thinking in the field.}

In the future our primary target will be the design of a more accurate action tube detector with the ability to perform better in challenging scenarios such as those portrayed in ROAD or SARAS-ESAD. 
We will also explore the end-to-end training of the entire model in all its three components.
Further down the line, 
we will update our S/T scene graph approach to properly model the heterogenous nature of the graph \cite{zhang2019heterogeneous}, 
and extend it to a more complete representation of complex dynamic events
in which nodes (rather than correspond all to action tubes) may be associated with any relevant elements of a dynamic scene, such as objects, agents, actions, locations and their attributes (e.g. red, fast, drivable, etc). 
\vspace{-5mm}

\section{Conclusions}

In this paper we presented a spatiotemporal complex activity detection framework which leverages both part deformation and a heterogenous graph representation. Our approach is based on three building blocks; action tube detection, part-based deformable 3D RoI pooling for feature extraction and a GCN module which processes the variable number of detected action tubes to model the overall semantics of a complex activity.
In an additional contribution, we temporally annotated two recently released benchmark datasets (ROAD and ESAD) in terms of {long-term} complex activities. Both datasets come with video-level action tube annotation, making them suitable benchmarks for future work in this area. We thoroughly evaluated our method, 
showing the effectiveness of our 3D part-based deformable model approach for the detection of complex activities.

\section*{Acknowledgements}
The work reported in this paper was supported by Huawei Technologies Co., Ltd. and the European Union’s Horizon 2020 research and innovation programme, under Grant Agreement no. 779813 (SARAS).

\bibliography{egbib}
\end{document}


\maketitle
\vspace{-10mm}



\section{Additional Details}
\label{sec:intro}

\subsection{Overview}
\begin{figure}[h]
    \centering
    \includegraphics[width=0.60\textwidth]{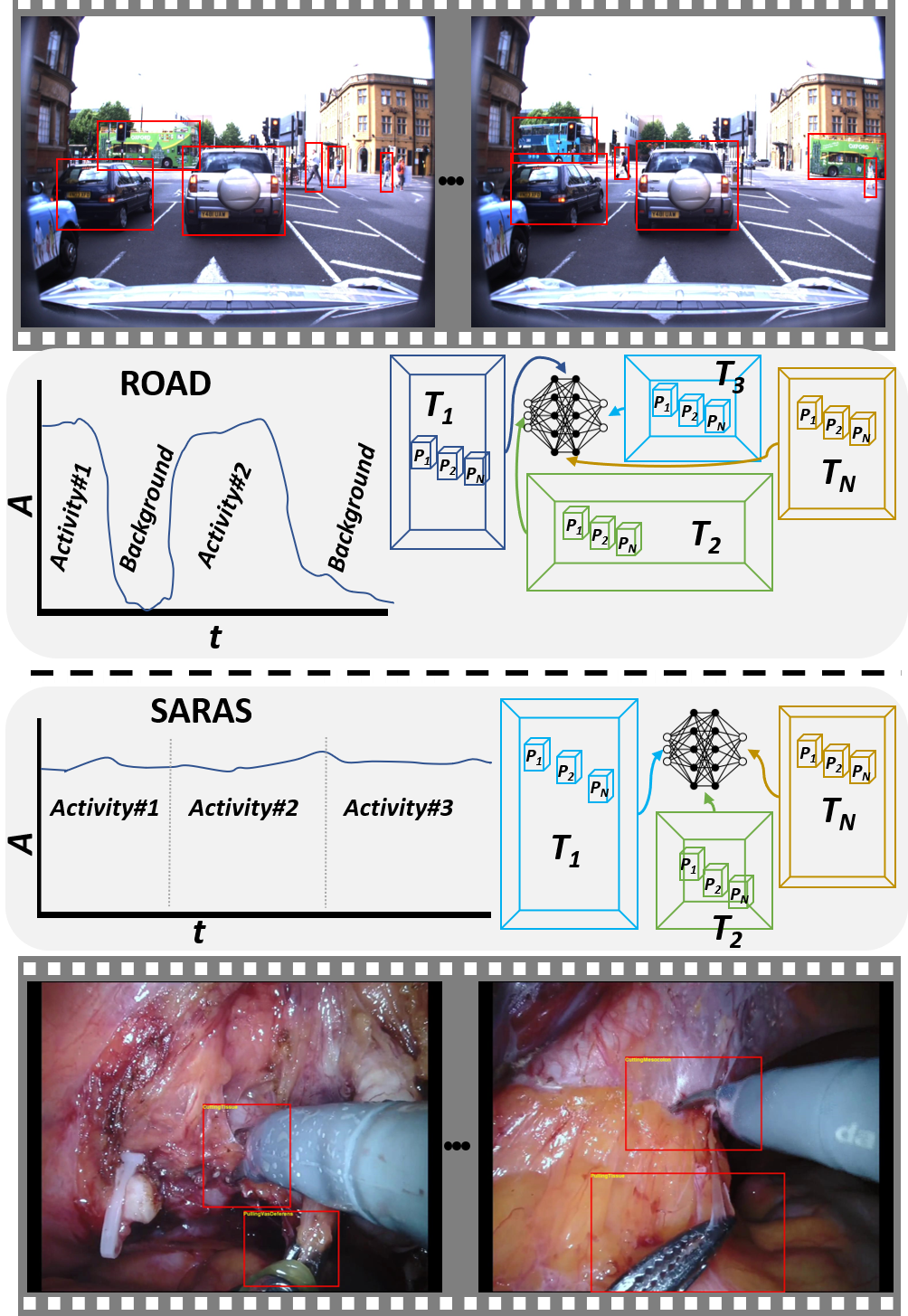}
    \caption{Our complex activity recognition framework workflow for two videos from the ROAD (autonomous driving) and SARAS (surgical robotics) datasets. The overall concept of detecting complex activities via a part-based deformation mechanism is the same in both cases. However, in ROAD background frames exist which do not contain any complex road activity. In contrast, the SARAS complex activities are actually the phases of a surgical procedure, which are contiguous without the need for a background label. 
    }
    \label{fig:workflow} \vspace{-3mm}
\end{figure}

A general overview of our complex activity framework is visualized in Figure \ref{fig:workflow}. 

\subsection{Tube Detection}
\begin{figure*}[h]
    \centering
    \includegraphics[width=0.75\textwidth]{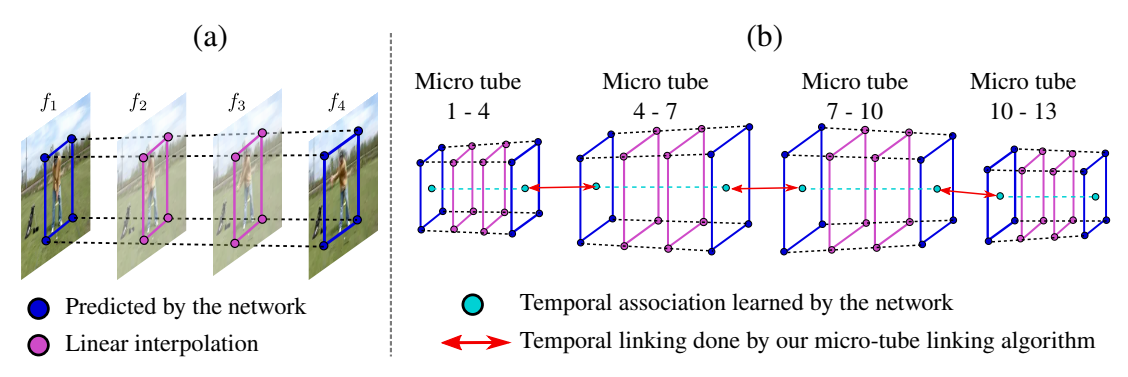}
    \caption{(a)
    AMTNet generates micro-action tubes  
    by predicting bounding boxes for the start and the end frame of a snippet. Predictions for a fixed number of
    intermediate frames are constructed by bilinear interpolation. (b) Complete action tubes are created by temporally linking micro action tubes by dynamic programming.}
    \label{fig:AMTNet}
\end{figure*}

The detail of AMTNet tube generation is illustrated in Figure \ref{fig:AMTNet}. 

\subsection{Algorithm}
The complete process of our framework is described in Algorithm \ref{algo:algo1}.

\begin{algorithm}[h]
\SetAlgoLined
\nl \KwIn{Input Video $V$ divided into $N$ number of Snippets  $S$ consists of fixed $M$ number of frames ${F}^{1,2,3,...,M}$}
\nl {\textbf{Initialization}: Load pretrained action tube detector ${ATD}$ on desire dataset, import our 3D deformable RoI pooling ${DRoI}$ and sparsity ${Sp}$ modules \; }
\nl \For{$i$ $\in$ 1,2,3,..., $N$}{
\nl  Extract features and detect action tubes ${B_A}^{M \times K \times 5}$, ${X_B}^{M \times 64 \times 300 \times 300}$  $\leftarrow$ ${ATD}$($S_i$) \;
\nl ${X_{DRoI}}^{K \times 64 \times M \times 7 \times 7}$ $\leftarrow$ ${DRoI({{B_A}^{M \times K \times 5}} ,{ {X_B}^{M \times 64 \times 300 \times 300} }  )}$ \;
\nl ${graphs}^k$ $\leftarrow$ ${Construct\_Graph}({X_{DRoI}}^{K \times 64 \times M \times 7 \times 7})$\;
\nl ${X_{G}}^{k \times 2048}$ $\leftarrow$ ${G}({graphs}^k)$  \;
\nl ${CA}_i$ $\leftarrow$ ${Softmax}({X_{G}}^{k \times 2048})$  \;
\nl  \eIf { ${CA}_{i-1}$ == ${CA}_{i}$}{
\nl   Same activity\;
    }{
\nl   \eIf{${CA}_{i-1}$ == ${CA}_{i+1}$}{
\nl   same activity\;
}{
\nl   end of activity\;
\nl   start new activity\;
}}}
\nl \KwOut{Start and end frames with activity labels}
 \caption{Complex Activity Detection}
 \label{algo:algo1}
\end{algorithm}

\section{Datasets Details}
Due to the lack of space here we provided the detail of both our augmented datasets with their statistics.
\begin{table}
\begin{center}
\footnotesize \scalebox{1} {
\begin{tabular}{p{0.016\textwidth}p{0.16\textwidth}p{0.06\textwidth}p{0.06\textwidth}p{0.50\textwidth}}
\hline
No & Complex activity & Train snippets & Test snippets & Scenario/Description \\
\hline\hline
1 & Negotiating intersection &  523 &  414 & Autonomous vehicle (AV) stops at intersection/junction, waits for other vehicles to pass, then resumes.   \\
2 & Negotiating pedestrian crossing & 381  & 25  & AV Stops at pedestrian crossing signal, waits for pedestrian(s) and cycles to cross, then resumes.  \\
3 & Waiting in a queue & 570  & 447  & AV stops and waits in a queue for a signal (either pedestrian crossing or junction traffic light). 
\\
4 & Merging into vehicle lane & 68  & 3  & AV stops and waits for bus/car to merge into the vehicle lane from the opposite side of the road.  \\
5 & Sudden appearance &  3 & 3  & A pedestrian/cycle/vehicle suddenly appears in front of the vehicle, coming from
a side street.  \\
6 & Walking in the middle of road & 13  &  19 & A pedestrian walking in the middle of the road in front of the AV, causing the AV to slow down or stop.  \\

\hline
\end{tabular}
}
\end{center}
\caption{List of ROAD complex activities with number of train and test snippets splits and brief description.}
\label{tab:tab1}
\end{table}

\begin{table}
\begin{center}
\footnotesize \scalebox{1} {
\begin{tabular}{p{0.014\textwidth}p{0.2\textwidth}p{0.030\textwidth}p{0.030\textwidth}p{0.50\textwidth}}
\hline
No & Complex activity/ Phase \# & Train snippets & Test snippets & Scenario/Description \\
\hline\hline
1 & Phase\#1 (Bladder mobilization) &  1808 & 999  & Tools only used for mobilization   \\
2 & Phase\#2 (AFP Dissection) &  6263 &  4455 & Anytime only Anterior Prostatic Fat (AFP) is removed  \\
3 & Phase\#3 (Bladder neck transection)& 8353  & 2419  & Whenever the scissor cut the neck, until all the neck is transected \\
4 & Phase\#4 (NVP dissection) &  361 & 46  & Located just under the seminal vescicles. \\
5 & Phase\#5 ( Exposure of seminal vescicles) & 2858  & 800  & White bags surrounding the bladder neck. Comprises fat removal.  \\
6 & Phase\#6 (Urethral division) & 2553  &  1685 & 	Start even when removing fat on the urethra. \\
7 & Phase\#7 (Prostate liberation)& 1092  & 2903  & 	After urethral division. It is not AFP. Stops only when completely free. \\
8 & Phase\#8 (Vesicourethral anastomosis)& 4430  & 4690  & 	Every time there is string and needle in the scene. Stops when urethra is linked to bladder\\
\hline
\end{tabular}
}
\end{center}
\caption{SARAS video phases with number of train and test snippets and a brief description.} 
\label{tab:tab2}
\end{table}

\subsection{ROAD} 
The detail of each class and its statistics with complete description is given in Table \ref{tab:tab1}.

\subsection{SARAS-ESAD}
The detail of each phase and its statistics with complete description is given in Table \ref{tab:tab2}.